\def\year{2022}\relax
\documentclass[letterpaper]{article} % DO NOT CHANGE THIS
\usepackage{aaai22}  % DO NOT CHANGE THIS
\usepackage{times}  % DO NOT CHANGE THIS
\usepackage{helvet}  % DO NOT CHANGE THIS
\usepackage{courier}  % DO NOT CHANGE THIS
\usepackage[hyphens]{url}  % DO NOT CHANGE THIS
\usepackage{graphicx} % DO NOT CHANGE THIS
\usepackage{natbib}  % DO NOT CHANGE THIS AND DO NOT ADD ANY OPTIONS TO IT
\usepackage{caption} % DO NOT CHANGE THIS AND DO NOT ADD ANY OPTIONS TO IT
\DeclareCaptionStyle{ruled}{labelfont=normalfont,labelsep=colon,strut=off} % DO NOT CHANGE THIS
\usepackage{algorithm}
\usepackage{algorithmic}
\usepackage{amsmath}
\usepackage{array}
\usepackage{booktabs}
\usepackage{multirow}
\usepackage{subcaption}
\usepackage{newfloat}
\usepackage{listings}
\floatstyle{ruled}
\newfloat{listing}{tb}{lst}{}
\floatname{listing}{Listing}
\title{Attention-based Transformation from Latent Features to Point Clouds}
\author{    
    Kaiyi Zhang\textsuperscript{\rm 1},
    Ximing Yang\textsuperscript{\rm 1},
    Yuan Wu\textsuperscript{\rm 1},
    Cheng Jin\textsuperscript{\rm 1,2}
}
\affiliations{
    \textsuperscript{\rm 1}School of Computer Science, Fudan University, Shanghai, China\\
    \textsuperscript{\rm 2}Peng Cheng Laboratory, Shenzhen, China\\
    \{zhangky20, xmyang19, wuyuan, jc\}@fudan.edu.cn
}

\usepackage{bibentry}
\begin{document}

% \linenumbers
\maketitle

\begin{abstract}
In point cloud generation and completion, 
previous methods for transforming latent features to point clouds are generally based on fully connected layers (FC-based) or folding operations (Folding-based). 
However, 
point clouds generated by FC-based methods are usually troubled by outliers and rough surfaces. 
For folding-based methods, their data flow is large, convergence speed is slow, and they are also hard to handle the generation of non-smooth surfaces. 
In this work, 
we propose AXform, an attention-based method to transform latent features to point clouds. 
AXform first generates points in an interim space, using a fully connected layer. 
These interim points are then aggregated to generate the target point cloud. 
AXform takes both parameter sharing and data flow into account, 
which makes it has fewer outliers, fewer network parameters, and a faster convergence speed. 
The points generated by AXform do not have the strong 2-manifold constraint, which improves the generation of non-smooth surfaces. 
When AXform is expanded to multiple branches for local generations, 
the centripetal constraint makes it has properties of self-clustering and space consistency, 
which further enables unsupervised semantic segmentation. 
We also adopt this scheme and design AXformNet for point cloud completion. 
Considerable experiments on different datasets show that our methods achieve state-of-the-art results.
\end{abstract}

\section{Introduction}

\begin{figure}[htb]
    \centering
    \includegraphics[width=\columnwidth]{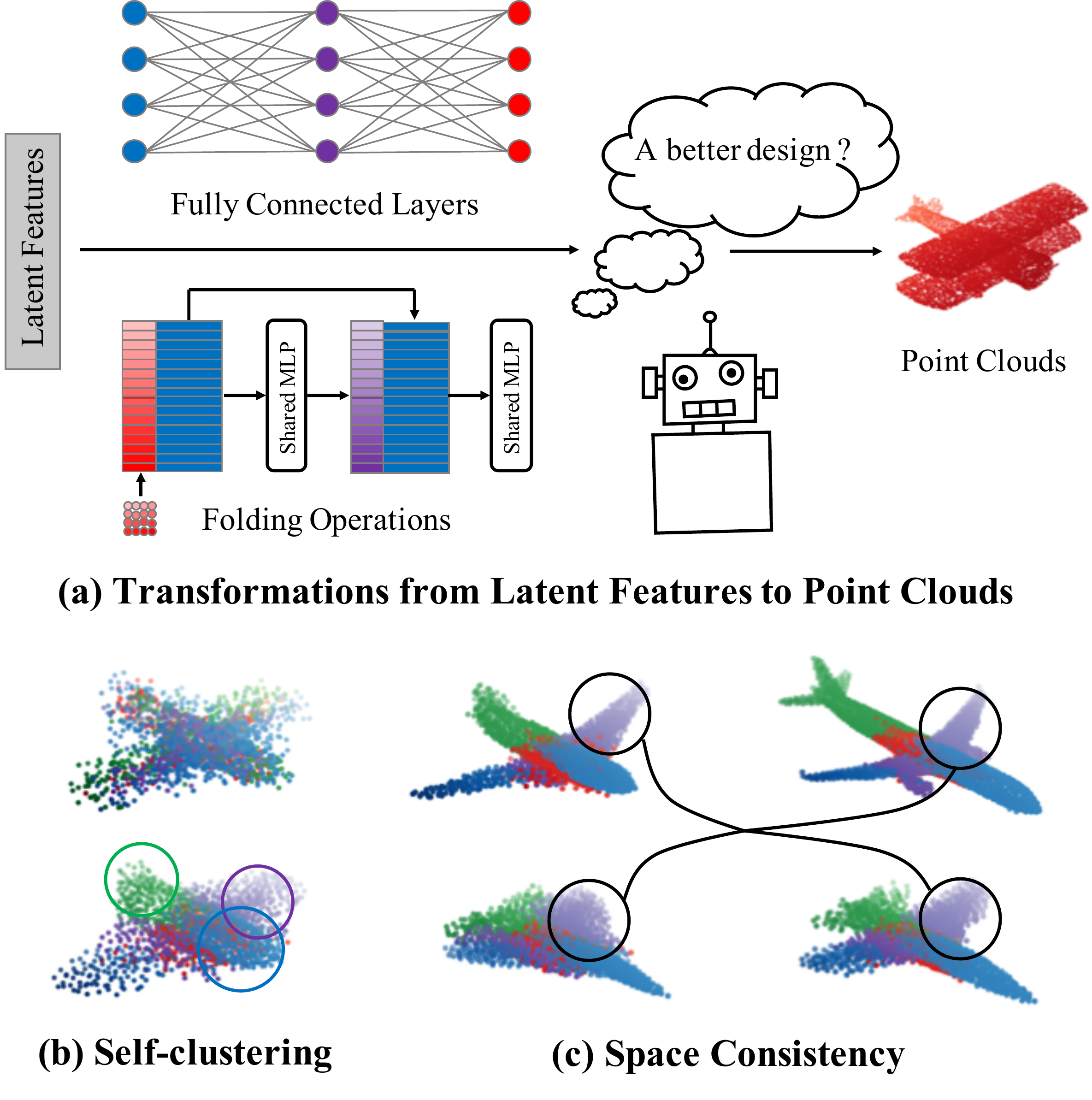}
    \caption{(a) Two commonly used methods for the transformation. 
    (b, c) Two properties of AXform with K branches.}
    \label{fig:introduction}
\end{figure}

% The point cloud has attracted more and more attention recent years. 
% The reason is that, 
% on the one hand, 
% popular devices such as Kinect and iPhone begin to embed RGB-D depth cameras, 
% creating rich data conditions for the development of the point cloud field; 
% on the other hand, 
% previously 3d object representation such as the distance fields, 
% the meshes and the voxel grids have disadvantages like high storage and high computational cost, 
% while point cloud can overcome these disadvantages well.

Recently, there have been considerable deep-learning-based point cloud generation and completion methods based on the architecture of autoencoder. 
Methods such as 
PointNet \cite{qi2016pointnet}, PointNet++ \cite{qi2017pointnetplusplus}, DGCNN \cite{dgcnn}, and Point Transformer \cite{zhao2020point} 
have studied encoder in detail to obtain better point cloud representations. 
There are also many methods exploring the design of the decoder. 
For example, \cite{Fan_2017_CVPR, achlioptas2017latent_pc, yuan2018pcn, Yang_Wu_Zhang_Jin_2021} simply use fully connected layers to generate coarse outputs. 
\cite{Yang_2018_CVPR, groueix2018, liu2019morphing, Wen_2020_CVPR} deform 2d squares into 3d surfaces. 
As shown in Figure \ref{fig:introduction}(a), 
fully connected layers and folding operations are two commonly used methods for the transformation from latent features to point clouds. 
However, they do not fully explore the transformation from aspects of point constraints, network parameters, data flow, and convergence speed.

As shown in the upper airplane in Figure \ref{fig:introduction}(b), 
point clouds generated by the FC-based method usually have outliers and rough surfaces. 
It is because the generated points do not share parameters, i.e., the available parameters of each point are few. 
When it is expanded to multiple branches, 
the generated point clouds do not have the property of self-clustering. 
The property of self-clustering means points generated by each branch are gathered together and it reflects the locality of an object. 
The lower airplane in Figure \ref{fig:introduction}(b) is an example of a self-clustered point cloud. 
The reason for this problem is that each generated point is free and not subject to a centripetal constraint.

Different from the FC-based method, 
the folding-based method has much fewer parameters and uses grid priors to constrain the generated points on a smooth 2-manifold. 
This constraint enables it to be expanded to multiple branches. 
However, its convergence speed is slow as it is difficult to deform a 2d square into various 3d surfaces with only a few parameters. 
In the same time, 3d objects often have non-smooth surfaces. 
Folding operations are also hard to handle these situations as the generated surfaces are folded from smooth 2-manifolds.

To address these issues, we propose an attention-based method AXform to transform latent features to point clouds. 
It first adopts a fully connected layer to generate points in an interim space. 
Then weighting operations are performed by an attention mechanism to aggregate target points in the interim space. 
Finally, these points are mapped to 3d space, which are the final point cloud. 
AXform has four advantages. 
First, it has fewer parameters comparing with FC-based and folding-based methods, which is benefit from sharing network parameters. 
Second, its convergence speed is faster than these two methods. 
Third, the points generated by AXform are more likely to concentrate due to a centripetal constraint. 
Thus AXform can be easily expanded to multiple branches for local generations. 
Last, as shown in Figure \ref{fig:introduction}(c), 
our AXform with K branches has a property of space consistency, 
which further enables unsupervised semantic segmentation.

Our main contributions are the following:
\begin{itemize}
    \item We propose a novel attention-based method for the transformation from latent features to point clouds. 
    Considerable experiments show that it performs better than previous methods.
    \item The proposed AXform can be expanded to multiple branches for local generations. 
    It has the properties of self-clustering and space consistency, 
    which can be used for unsupervised semantic segmentation on the generated point clouds.
    \item We apply AXform to point cloud completion, and it achieves state-of-the-art results on the PCN dataset.
\end{itemize}

\section{Related Work}

\textbf{Attention in Point Clouds} Attention mechanism has been widely used in point clouds to get better point cloud representations. 
\cite{lee2019set} presents an attention-based network to simulate interactions between elements in point clouds. 
\cite{Yang_2019_CVPR} uses attention layers to capture the relationship between points. 
\cite{li2019pugan} introduces a self-attention unit to enhance the quality of feature integration when upsampling point clouds. 
\cite{zhang2020eccv} uses an attention module to optimize the input point cloud, which reduces outliers in the generated point cloud. 
\cite{Wen_2020_CVPR} proposes a skip-attention model to capture the geometric information from local regions of the input to get a better representation.
There are also many other works exploring the application of attention in point clouds 
\cite{fuchs2020se3transformers, 8897042, liu2019l2gautoencoder, hu2019randla, ijcai2020-110}.

\noindent \textbf{Point Cloud Generation} Plenty of tasks need to generate point clouds as the output, which attracted a lot of research interest.
\cite{Fan_2017_CVPR} generates point clouds by using a fully connected branch and a 2d deconvolution branch. 
\cite{Yang_2018_CVPR} proposes an idea of deforming a 2d square into a 3d surface called folding, which has fewer parameters. 
\cite{groueix2018} further expands the folding operation to multiple branches and achieves better surface reconstructions. 
\cite{sun2018pointgrow} presents a novel autoregressive model by generating points one by one like 3D printing. 
\cite{Yang_Wu_Zhang_Jin_2021} generates a structural skeleton to achieve controllable generation. 
\cite{valsesia2018learning, hui2020pdgn} design deconvolution operations on point clouds to do generation. 
\cite{achlioptas2017latent_pc, li2018point, zamorski2018adversarial, Shu_2019_ICCV} proposes some GAN models. 
\cite{Yang_2019_ICCV, klokov20eccv, NEURIPS2020_bdbca288, luo2021diffusion} explore flow-based models for reversible point cloud generation.

\noindent \textbf{Point Cloud Completion} This task aims to complete authentic point clouds when given inputs with various missing patterns. 
It can contribute to a series of downstream applications like robotics operations \cite{varley2017shape}, 
scene understanding \cite{dai2018scancomplete}, 
and virtual operations of complete shapes \cite{visapp17}. 

Some methods explore to apply deep learning on supervised point cloud completion. 
\cite{yuan2018pcn} provides an autoencoder to combine the global and local shape information for point cloud completion. 
\cite{liu2019morphing} generates patches by using multiple branches to get a better locality. 
\cite{xie2020grnet} transforms point cloud into a new voxel representation, which better retains the spatial information of original partial point clouds. 
\cite{Xie_2021_CVPR} designs a differentiable renderer and applies adversarial training to realize better point supervisions. 
Different from previous methods, 
\cite{wen2021pmp} first proposes the idea of completing the objects by moving original points step by step. 
In addition, 
\cite{topnet2019, Hu_2019_ICCV, Sarmad_2019_CVPR, huang2020pfnet, DBLP:journals/corr/abs-2008-07358, zhang2020eccv, Wen_2020_CVPR, Wang_2020_CVPR, 9093117, Alliegro_2021_CVPR, pan2021variational} 
play an important role in promoting point cloud completion.

% There also exist methods for weak or unsupervised point cloud completion. 
% \cite{Stutz_2018_CVPR} is one of the pioneering works. 
% It directly measures the maximum likelihood between the latent representation of incomplete and complete shapes. 
% \cite{chen2020pcl2pcl} designs a GAN network to realize the training without requiring the correspondence complete point clouds, 
% which is more like a point cloud generation while keeping the partial unchanged. 
% \cite{wu_2020_ECCV, wen2021c4c, Zhang_2021_CVPR} further expand the GAN architecture to Conditional GAN, CycleGAN, and GAN Inversion respectively.

\section{Approach}

In this section, we first describe the framework of our method AXform and compare it with previous methods. 
Then we expand AXform to multiple branches and compare it with AtlasNet \cite{groueix2018}. 
Finally, we show how to apply AXform to point cloud completion.

\subsection{Our Method}

\begin{figure*}[htb]
    \centering
    \includegraphics[width=\textwidth]{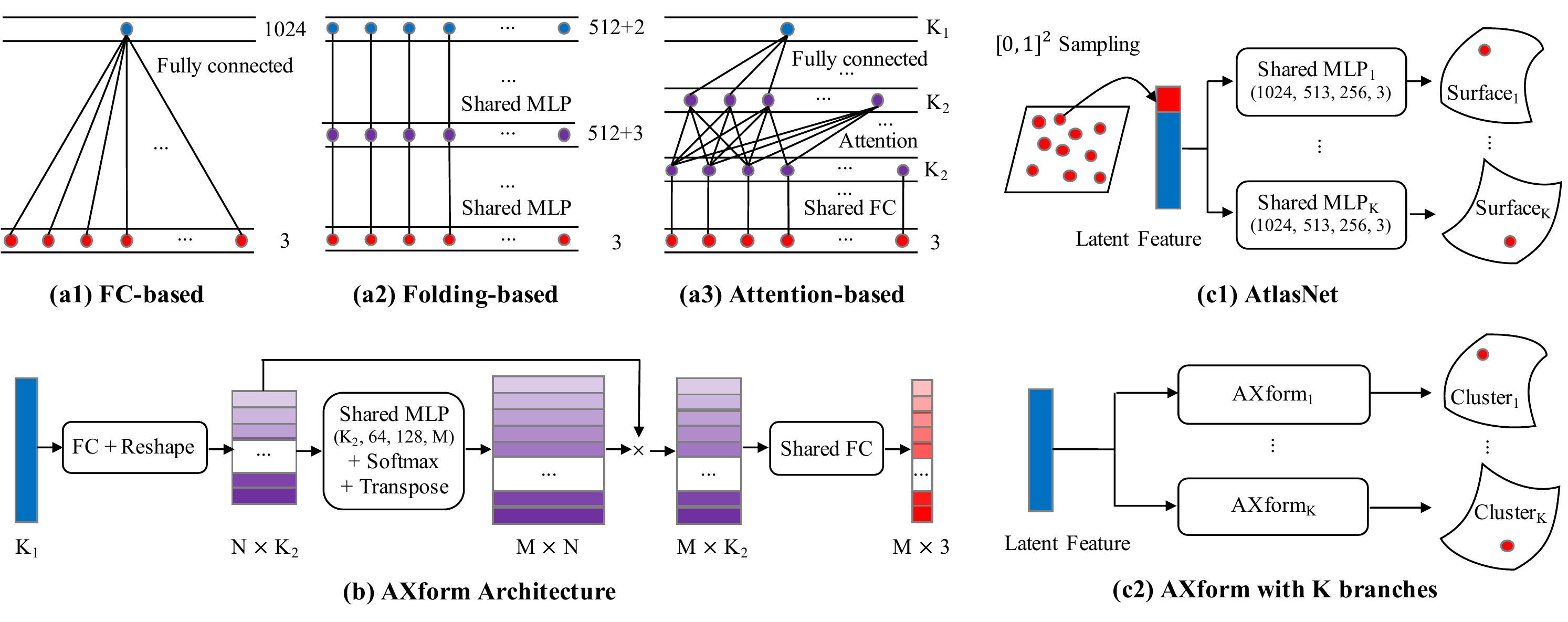}
    \caption{(a-1,2,3) The architectural difference between AXform and previous FC-based and folding-based methods. 
    The number on the right of each figure represents the space dimension. 
    (b) The architecture of AXform with one branch. 
    (c-1,2) The architectural difference between AtlasNet and our AXform with K branches.}
    \label{fig:our_method}
\end{figure*}

\subsubsection{AXform Framework}

AXform aims to achive a better transformation from final latent features to point clouds. 
As shown in Figure \ref{fig:our_method}(b), 
it includes three steps: interim points generation by fully connected layers, attention-based interim points aggregation, and 3D-mapping by shared FC.

Assuming that the input features $f_{in} \in R^{K_1}$. AXform first uses a fully connected layer to generate $N$ points in $R^{K_2}$ from $f_{in}$.
From these $N$ points, an attention mechanism is adopted to generate an $M * N$ attention map, where $M$ is the final number of points we want to generate.
Through this attention map, AXform aggregates these $N$ interim points to $M$ new points. 
The aggregation process is a convex combination process,
therefore the $M$ points are ensured to be in the convex hull constructed by the $N$ interim points in $R^{K_2}$, 
thus the locality of the $M$ points are guaranteed. 
In detail, 
a shared MLP $(K_2, 64, 128, M)$ are firstly adopted to transform $N$ points in $R^{K_2}$ into an $N \times M$ matrix. 
Then softmax activation function is applied on the dimension of $N$ and the matrix is transposed to form an $M \times N$ attention map. 
Finally, the original $N$ points are aggregated by the attention map to generate $M$ new points in $R^{K_2}$.
After the attention-based interim points aggregation, 
each point of the $M$ points is mapped to a 3d point by a shared fully connected layer.

\subsubsection{Comparison with FC and Folding}

Fully connected layers and folding operations are two commonly used methods for the transformation from latent features to point clouds. 
As shown in Figure \ref{fig:our_method}(a-1,2,3), 
we regard features as points in a high-dimensional space and mark them blue. 
Intermediate feature points are marked purple and final outputs are marked red. 

For the FC-based method, 
each generated point only uses about $1024 \times 3$ parameters. 
Instead, points generated by AXform share the first fully connected layer $(K_1 \rightarrow K_2)$. 
Combining Figure \ref{fig:our_method}(b), 
we can easily infer that each point generated by AXform uses about $K_1 \times K_2 \times N$ parameters. 
In our experiments, 
we set $K_1 = 128, K_2 = 32, N = 128$, and thus $K_1 \times K_2 \times N$ is much larger than $1024 \times 3$.
Each point use more parameters attentively 
can not only improves the quality of outputs but also make the outputs have a property of self-clustering. 
In addition, 
the number of whole parameters in AXform is mainly contributed by the first fully connected layer $(K_1 \times K_2 \times N)$. 
It is much fewer than what in the FC-based method $(1024 \times 2048 \times 3)$.

For the folding-based method, 
since each point in a 2d grid is concatenated by the same 512-dim latent feature, 
the data flow in the network is large. 
Its scale depends on the number of target points. 
Instead, the main memory consumption of data flow in AXform lies in the multiplication of the attention map and the interim point set. 
It depends on the number of interim points. 
By setting $N < M$, AXform generally takes up less memory. 
In addition, the folding-based method uses grid priors to constrain the generated points on a smooth 2-manifold. 
However, this prior is often too strong, leading to a slow convergence speed. 
But AXform has a weaker constraint on the generated points based on the attention mechanism, so the convergence speed is quite faster.

\subsection{Multiple Branches}

Since AXform uses an attention mechanism to generate each target point, 
and its weighting operation can be regarded as a convex combination. 
Assuming that the convex hull formed by the interim point set $S$ containing $N$ points is $conv(S)$, 
the final output containing $M$ points will fall inside $conv(S)$. 
This centripetal constraint forces the final output to be more concentrated rather than scattered. 
Therefore, when we expand AXform to multiple branches, 
points generated by each branch will have a property of self-clustering. 
Based on this property and referring to the AtlasNet \cite{groueix2018} architecture, 
we propose AXform with K branches, as shown in Figure \ref{fig:our_method}(c2).

Different from AtlasNet which combines each point sampled from a 2d square with the same latent feature and then performs multi-branch folding, 
our method directly transforms the latent feature into multiple small point clusters by using AXforms. 
In the following experiments, 
we will compare them in detail to show the superiority of our method.

\begin{figure*}[htb]
    \centering
    \includegraphics[width=\textwidth]{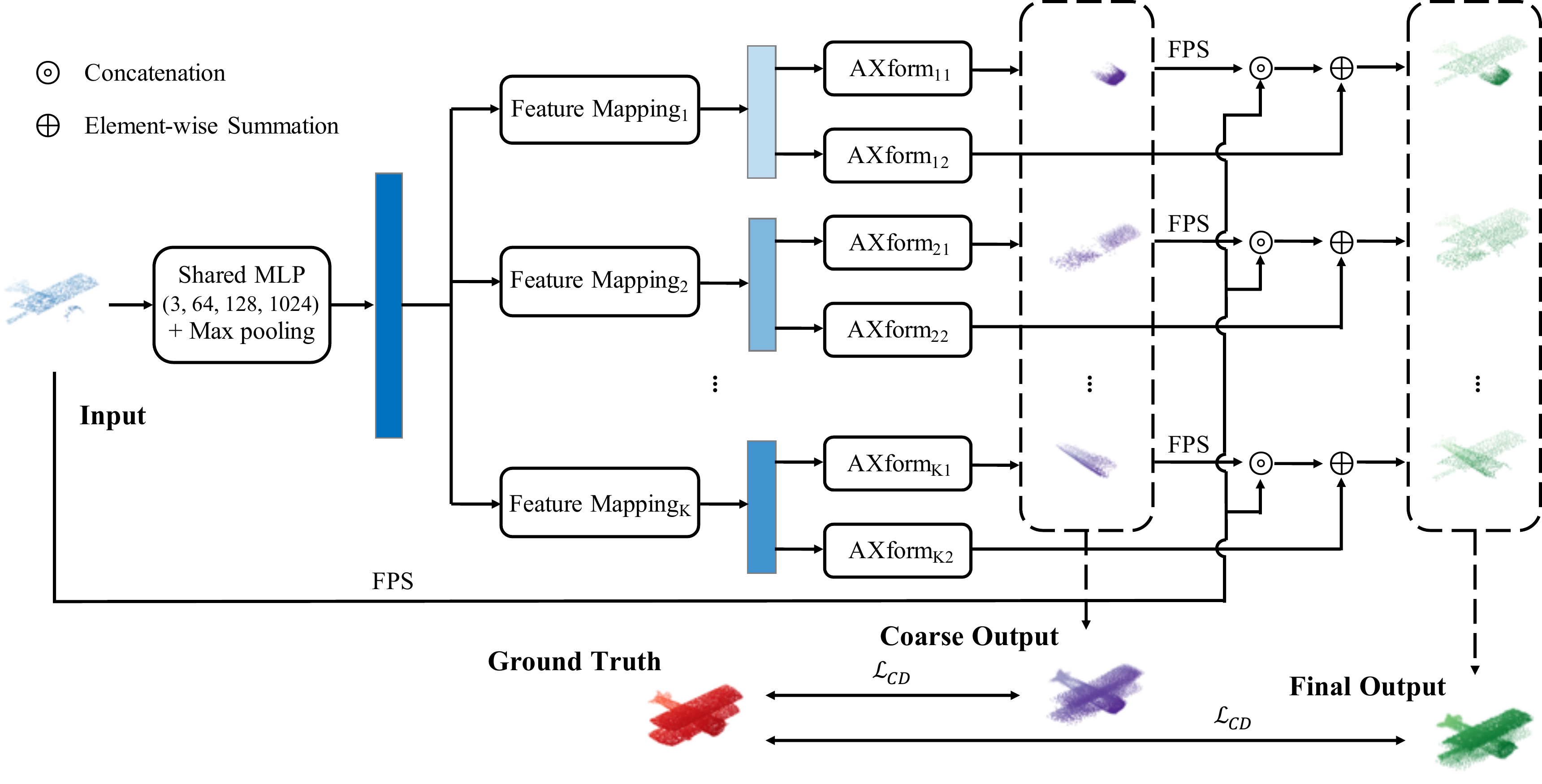}
    \caption{The architecture of AXformNet. It includes K branches and two stages. 
    Each branch generates a part of the target point cloud. 
    The first stage uses AXform to generate a coarse point cloud. 
    In the second stage, refinement is performed by combining input to generate a final point cloud.}
    \label{fig:apply_to_pcc}
\end{figure*}

\subsection{Application on Point Cloud Completion}

Since previous methods have done considerable exploration on the use of autoencoder for point cloud completion, 
we propose a network based on AXform for supervised point cloud completion, called AXformNet.

\subsubsection{AXformNet Architecture}

As shown in Figure \ref{fig:apply_to_pcc}, 
the input of AXformNet is partial point clouds generated by back-projecting 2.5D depth images into 3D 
and the output is completed point clouds. 
The framework of AXformNet is based on an autoencoder. 
To better demonstrate the superiority of our decoder, 
we simply design our encoder with a shared MLP layer $(3, 64, 128, 1024)$ and a max pooling layer, 
which is like what in PointNet \cite{qi2016pointnet}. 
The design of our decoder is based on AXform with K branches. 
The difference is that each branch contains two AXforms and an additional feature mapping module.

Since different branches focus on the generation of different parts of an object, 
the input feature of AXform in each branch should be different. 
Therefore, 
we add a feature mapping module before each branch to achieve adaptive feature transformation, 
which is a 4-layer fully connected network $(1024, 1024, 1024, 128)$.

Following the idea of considerable previous methods like \cite{liu2019morphing, Wang_2020_CVPR}, 
we also add a refinement network after the generation of coarse outputs. 
Specifically, 
the coarse output of each branch and the input partial point cloud are first concatenated after farthest point sampling (FPS) respectively. 
Then a bias calculated by AXform is added to generate the final output. 
Here we need to point out that for a fair comparison, 
we also combine the partial point cloud when performing the refinement. 
This operation contributes to the metric of Chamfer Distance a lot. 
Assuming that a partial point cloud occupies $\lambda$ of its corresponding ground truth point cloud in space, 
and their points are completely coincident. 
If $CD(complete, gt) = A$ and we replace a part of the complete point cloud with the original partial point cloud to get $complete'$, 
it is easy to infer that $CD(complete', gt) = (1-\lambda)A$, which is a large improvement of the metric.

\subsubsection{Loss Functions}
We use L1 Chamfer Distance \cite{yuan2018pcn} to supervise the training process of AXformNet. 
$Y_{coarse}$, $Y_{final}$ and $Y_{gt}$ represent the coarse output point cloud, the final output point cloud, and the ground truth point cloud respectively. 
$\alpha$ is a weighting factor. 
The total loss for training is then given as: 
\begin{equation}
    \mathcal{L} = \alpha \mathcal{L}_{CD} (Y_{coarse}, Y_{gt}) + \mathcal{L}_{CD} (Y_{final}, Y_{gt})
\end{equation}

\section{Experiments}

\subsection{Datasets and Implementation Details}

We evaluate AXform on three representative categories in the ShapeNet \cite{Chang:2015:SAI} dataset: $airplane$, $car$, and $chair$. 
The point clouds are obtained by sampling points uniformly from the mesh surface. 
All the point clouds are then centered and scaled to [-0.5, 0.5]. 
We follow the train/val/test split in ShapeNet official documents and use 2048 points for each shape during both training and testing. 
A simple shared MLP layer $(3, 64, 128, K_1)$ with a max pooling layer is used as the encoder. 
All the experiments are performed for 200 epochs with a batch size of 32. 
Adam is used as the optimizer and the initial learning rate is 1e-4. 
The Chamfer Distance used in these experiments is L2 Chamfer Distance \cite{Fan_2017_CVPR}.

We also evaluate AXformNet on the PCN \cite{yuan2018pcn} dataset for point cloud completion. 
It includes 30974 shapes with 8 categories. 
We set branch number $K = 16$ and train our method for 100 epochs with a batch size of 128. 
$\alpha$ increases from 0.01 to 1 within the first 25 epochs. 
Adam is used as the optimizer and the initial learning rate is 1e-3.

\subsection{Reconstruction and Generation}

\begin{table}[htb]
    \centering
    \caption{Quantitative comparison of reconstruction results on our sampled ShapeNet dataset. 
    CD is multiplied by $10^4$.}
    \label{tab:reconstruction}
    \resizebox{\columnwidth}{!}{
    \begin{tabular}{c c c *3c *2c}
        \toprule
        \#Branches & Category & Methods & $K_1$ & $K_2$ & $N$ & CD $\downarrow$ & Params. $\downarrow$ \\
        \midrule
        \multirow{13}{*}{K = 1} & \multirow{3}{*}{Airplane} & FC-based & 1024 & & & 7.895 & 7.4M \\
        & & Folding-based & 512 & & & 9.208 & 1.1M \\
        & & Ours & 128 & 32 & 128 & \textbf{4.386} & \textbf{0.8M} \\
        \cmidrule(lr){2-8}
        & \multirow{3}{*}{Car} & FC-based & 1024 & & & 11.523 & 7.4M \\
        & & Folding-based & 512 & & & 20.989 & 1.1M \\
        & & Ours & 128 & 32 & 128 & \textbf{8.008} & \textbf{0.8M} \\
        \cmidrule(lr){2-8}
        & \multirow{3}{*}{Chair} & FC-based & 1024 & & & 13.861 & 7.4M \\
        & & Folding-based & 512 & & & 23.103 & 1.1M \\
        & & Ours & 128 & 32 & 128 & \textbf{11.606} & \textbf{0.8M} \\
        \cmidrule(lr){2-8}
        & \multirow{3}{*}{All} & FC-based & 1024 & & & 8.578 & 7.4M \\
        & & Folding-based & 512 & & & 12.980 & 1.1M \\
        & & Ours & 128 & 32 & 128 & \textbf{8.046} & \textbf{0.8M} \\
        \midrule
        \multirow{9}{*}{K = 16} & \multirow{2}{*}{Airplane} & AtlasNet & 1024 & & & 6.307 & 27.5M \\
        & & Ours & 128 & 32 & 128 & \textbf{3.607} & \textbf{8.9M} \\
        \cmidrule(lr){2-8}
        & \multirow{2}{*}{Car} & AtlasNet & 1024 & & & 15.269 & 27.5M \\
        & & Ours & 128 & 32 & 128 & \textbf{7.670} & \textbf{8.9M} \\
        \cmidrule(lr){2-8}
        & \multirow{2}{*}{Chair} & AtlasNet & 1024 & & & 17.154 & 27.5M \\
        & & Ours & 128 & 32 & 128 & \textbf{9.423} & \textbf{8.9M} \\
        \cmidrule(lr){2-8}
        & \multirow{2}{*}{All} & AtlasNet & 1024 & & & 11.057 & 27.5M \\
        & & Ours & 128 & 32 & 128 & \textbf{6.867} & \textbf{8.9M} \\
        \bottomrule
    \end{tabular}}
\end{table}

\begin{figure}[htb]
    \centering
    \includegraphics[width=\columnwidth]{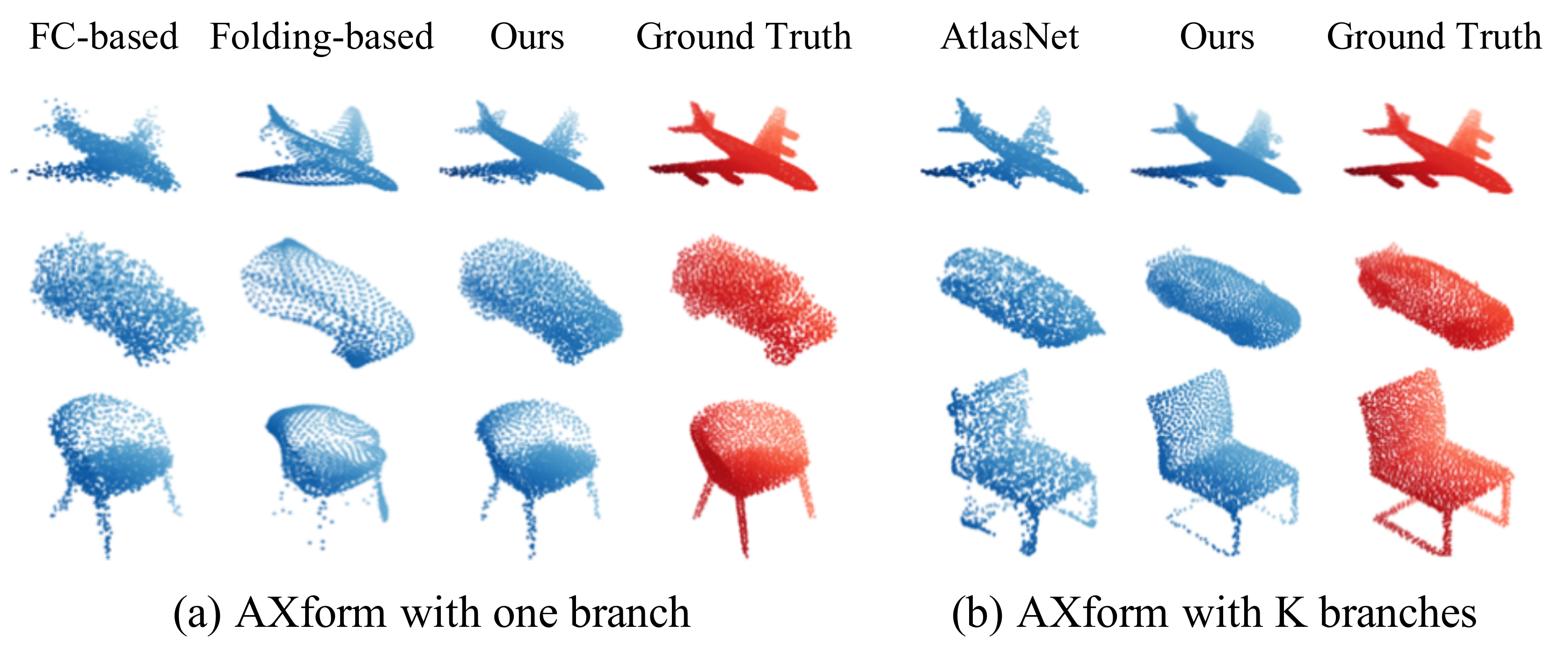}
    \caption{Visualized comparison of reconstruction results on our sampled ShapeNet dataset. 
    Trained on each category.}
    \label{fig:recon_comparison}
\end{figure}

\begin{figure}[htb]
    \centering
    \includegraphics[width=\columnwidth]{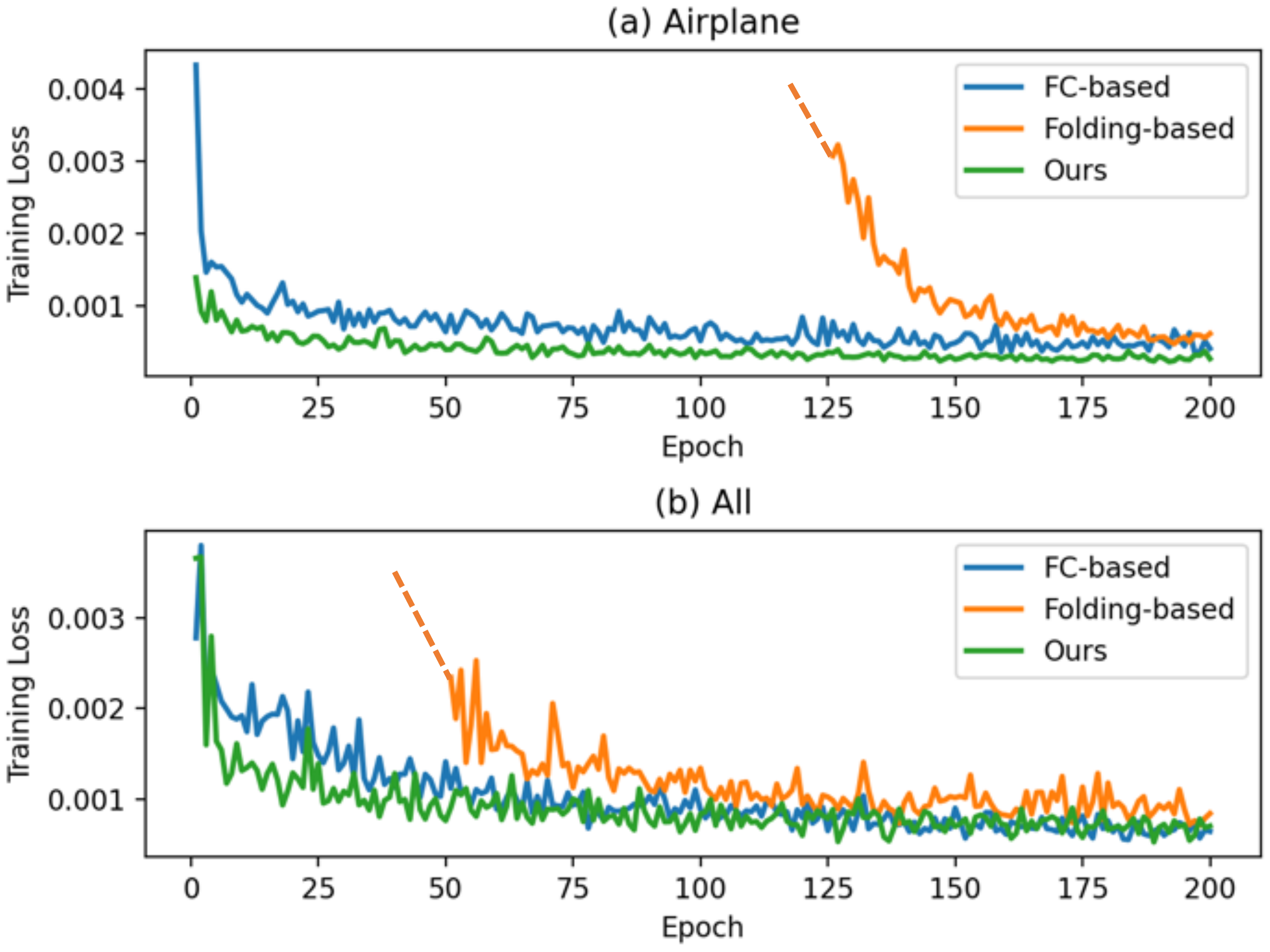}
    \caption{Loss curve during training.}
    \label{fig:recon_loss}
\end{figure}

\begin{table}[htb]
    \centering
    \caption{Ablation studies for the branch number K. $K_1=128$ and $K_2=32$ are fixed. CD is multiplied by $10^4$. Trained on airplane category.}
    \label{tab:branches_ablation}
    \resizebox{\columnwidth}{!}{
    \begin{tabular}{c c *5c}
        \toprule
        & \#Branches & 2 & 4 & 8 & 16 & 32 \\
        \cmidrule(lr){2-7}
        \multirow{2}{*}{Fix $N=128$} & CD & 4.280 & 4.041 & 3.840 & 3.607 & 3.520 \\
        & Params. & 1.3M & 2.4M & 4.6M & 8.9M & 17.5M \\
        \midrule
        \multirow{2}{*}{Fix Params. $\approx$ 8.9M} & CD & 3.844 & 3.774 & 3.724 & 3.607 & 3.643 \\
        & $N$ & 1024 & 512 & 256 & 128 & 64 \\
        \bottomrule
    \end{tabular}}
\end{table}

\begin{table}[htb]
    \centering
    \caption{Quantitative comparison of replacing the MLP decoder in l-GAN (CD) with our AXform. 
    JSD, MMD-CD/EMD are multiplied by $10^2$, $10^2$, and $10^3$ respectively. Trained on airplane category.}
    \label{tab:generation}
    \resizebox{\columnwidth}{!}{
    \begin{tabular}{c *7c}
        \toprule
        \multirow{2}{*}{Methods} & \multirow{2}{*}{JSD $\downarrow$} & \multicolumn{2}{c}{MMD $\downarrow$} & \multicolumn{2}{c}{COV \%, $\uparrow$} & \multicolumn{2}{c}{1-NNA \%, $\downarrow$}  \\
        \cmidrule(lr){3-4} \cmidrule(lr){5-6} \cmidrule(lr){7-8}
        & & CD & EMD & CD & EMD & CD & EMD \\
        \midrule
        l-GAN (CD) & 7.24 & \textbf{0.454} & \textbf{4.43} & \textbf{33.66} & \textbf{25.70} & 63.00 & 81.23 \\
        l-GAN-AXform & \textbf{6.27} & 0.498 & 4.54 & 33.33 & 24.59 & \textbf{61.59} & \textbf{78.55} \\
        \bottomrule
    \end{tabular}}
\end{table}

\begin{figure}[htb]
    \centering
    \includegraphics[width=\columnwidth]{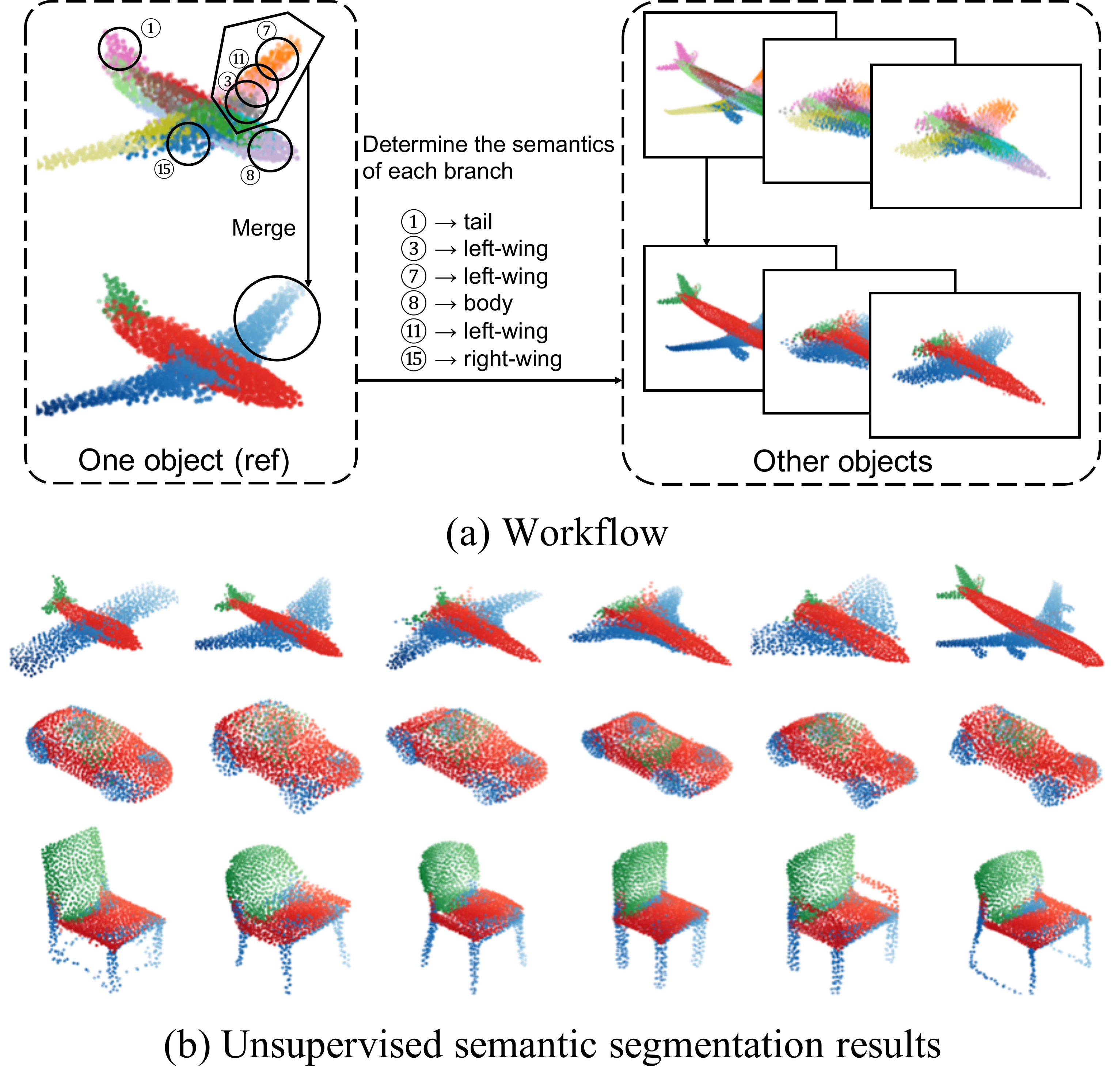}
    \caption{AXform with K branches has a property of space consistency which enables unsupervised semantic segmentation on the generated point clouds.}
    \label{fig:uss_results}
\end{figure}

\begin{table*}[htb]
    \scriptsize
    \centering
    \setlength\tabcolsep{1pt}
    \caption{Quantitative comparison between our methods and existing methods. The metric is CD, multiplied by $10^3$.}
    \label{tab:pcc_comparison_cd}
    %\resizebox{\columnwidth}{!}{
    \begin{tabular}{c | *8c | c | c | *8c | c}
        \toprule
        Methods & Airplane & Cabinet & Car & Chair & Lamp & Couch & Table & Watertcraft & Average & Methods & Airplane & Cabinet & Car & Chair & Lamp & Couch & Table & Watertcraft & Average \\
        \midrule
        FoldingNet & 9.491 & 15.796 & 12.611 & 15.545 & 16.413 & 15.969 & 13.649 & 14.987 & 14.308 & MSN & 5.596 & 11.963 & 10.776 & 10.620 & 10.712 & 11.895 & 8.704 & 9.485 & 9.969 \\
        AtlasNet & 6.366 & 11.943 & 10.105 & 12.063 & 12.369 & 12.990 & 10.331 & 10.607 & 10.847 & GRNet & 6.450 & 10.373 & 9.447 & 9.408 & 7.955 & 10.512 & 8.444 & 8.039 & 8.828 \\
        PCN & 5.502 & \textbf{10.625} & 8.696 & 10.998 & 11.339 & \textbf{11.676} & \textbf{8.590} & 9.665 & 9.636 & SpareNet & 5.956 & 12.567 & 9.956 & 11.931 & 11.105 & 13.388 & 9.950 & 9.589 & 10.555 \\
        TopNet & 7.614 & 13.311 & 10.898 & 13.823 & 14.439 & 14.779 & 11.224 & 11.124 & 12.151 & PMP-Net & 5.650 & 11.240 & 9.640 & 9.510 & \textbf{6.950} & 10.830 & 8.720 & \textbf{7.250} & 8.730 \\
        \midrule
        Ours(vanilla) & \textbf{5.366} & 10.677 & \textbf{8.646} & \textbf{10.743} & \textbf{10.458} & 11.683 & 8.727 & \textbf{9.300} & \textbf{9.450} & Ours & \textbf{4.760} & \textbf{10.178} & \textbf{8.600} & \textbf{9.133} & 8.173 & \textbf{10.395} & \textbf{7.752} & 7.803 & \textbf{8.349} \\
        \bottomrule
    \end{tabular}%}
\end{table*}

\begin{table*}[htb]
    \scriptsize
    \centering
    \setlength\tabcolsep{1pt}
    \caption{Quantitative comparison between our methods and existing methods. The metric is F-Score@1\%.}
    \label{tab:pcc_comparison_f1}
    \begin{tabular}{c | *8c | c | c | *8c | c}
        \toprule
        Methods & Airplane & Cabinet & Car & Chair & Lamp & Couch & Table & Watertcraft & Average &Methods & Airplane & Cabinet & Car & Chair & Lamp & Couch & Table & Watertcraft & Average \\
        \midrule
        FoldingNet & 0.642 & 0.237 & 0.382 & 0.236 & 0.219 & 0.197 & 0.361 & 0.299 & 0.322 &MSN & 0.885 & \textbf{0.644} & 0.665 & 0.657 & 0.699 & 0.604 & 0.782 & 0.708 & 0.705 \\
        AtlasNet & 0.845 & 0.552 & 0.630 & 0.552 & 0.565 & 0.500 & 0.660 & 0.624 & 0.616 &GRNet & 0.843 & 0.618 & 0.682 & 0.673 & 0.761 & 0.605 & 0.751 & 0.750 & 0.708 \\ 
        PCN & 0.881 & \textbf{0.651} & 0.725 & 0.625 & 0.638 & \textbf{0.581} & \textbf{0.765} & 0.697 & 0.695 &SpareNet & 0.869 & 0.571 & 0.672 & 0.592 & 0.647 & 0.527 & 0.719 & 0.690 & 0.661 \\
        TopNet & 0.771 & 0.404 & 0.544 & 0.413 & 0.408 & 0.350 & 0.572 & 0.560 & 0.503 &PMP-Net & 0.860 & 0.495 & 0.570 & 0.600 & 0.778 & 0.516 & 0.639 & 0.742 & 0.650 \\
        \midrule
        Ours(vanilla) & \textbf{0.893} & 0.634 & \textbf{0.725} & \textbf{0.632} & \textbf{0.667} & 0.567 & 0.756 & \textbf{0.703} & \textbf{0.697} & Ours & \textbf{0.920} & 0.642 & \textbf{0.734} & \textbf{0.704} & \textbf{0.782} & \textbf{0.605} & \textbf{0.801} & \textbf{0.778} & \textbf{0.746} \\
        \bottomrule
    \end{tabular}
\end{table*}

\begin{table}[htb]
    \centering
    \setlength\tabcolsep{1pt}
    \caption{Ablation studies for the feature mapping module.}
    \label{tab:pcc_comparison_ablation}
    \begin{minipage}{\linewidth}
    \centering
    \resizebox{\columnwidth}{!}{
    \begin{tabular}{c | *8c | c}
        \toprule
        Methods & Airplane & Cabinet & Car & Chair & Lamp & Couch & Table & Watertcraft & Average \\
        \midrule
        Ours(w/o fm) & 4.984 & 10.319 & 8.750 & 9.726 & 8.845 & 10.746 & 8.097 & 8.237 & 8.713 \\
        Ours & \textbf{4.760} & \textbf{10.178} & \textbf{8.600} & \textbf{9.133} & \textbf{8.173} & \textbf{10.395} & \textbf{7.752} & \textbf{7.803} & \textbf{8.349} \\
        \bottomrule
    \end{tabular}}
    \subcaption{CD.}
    \end{minipage}
    \begin{minipage}{\linewidth}
    \centering
    \resizebox{\columnwidth}{!}{
    \begin{tabular}{c | *8c | c}
        \toprule
        Methods & Airplane & Cabinet & Car & Chair & Lamp & Couch & Table & Watertcraft & Average \\
        \midrule
        Ours(w/o fm) & 0.912 & 0.629 & 0.717 & 0.671 & 0.747 & 0.584 & 0.782 & 0.758 & 0.725 \\ 
        Ours & \textbf{0.920} & \textbf{0.642} & \textbf{0.734} & \textbf{0.704} & \textbf{0.782} & \textbf{0.605} & \textbf{0.801} & \textbf{0.778} & \textbf{0.746} \\
        \bottomrule
    \end{tabular}}
    \subcaption{F-Score@1\%.}
    \end{minipage}
\end{table}

\begin{figure*}[htb]
    \centering
    \includegraphics[width=\textwidth]{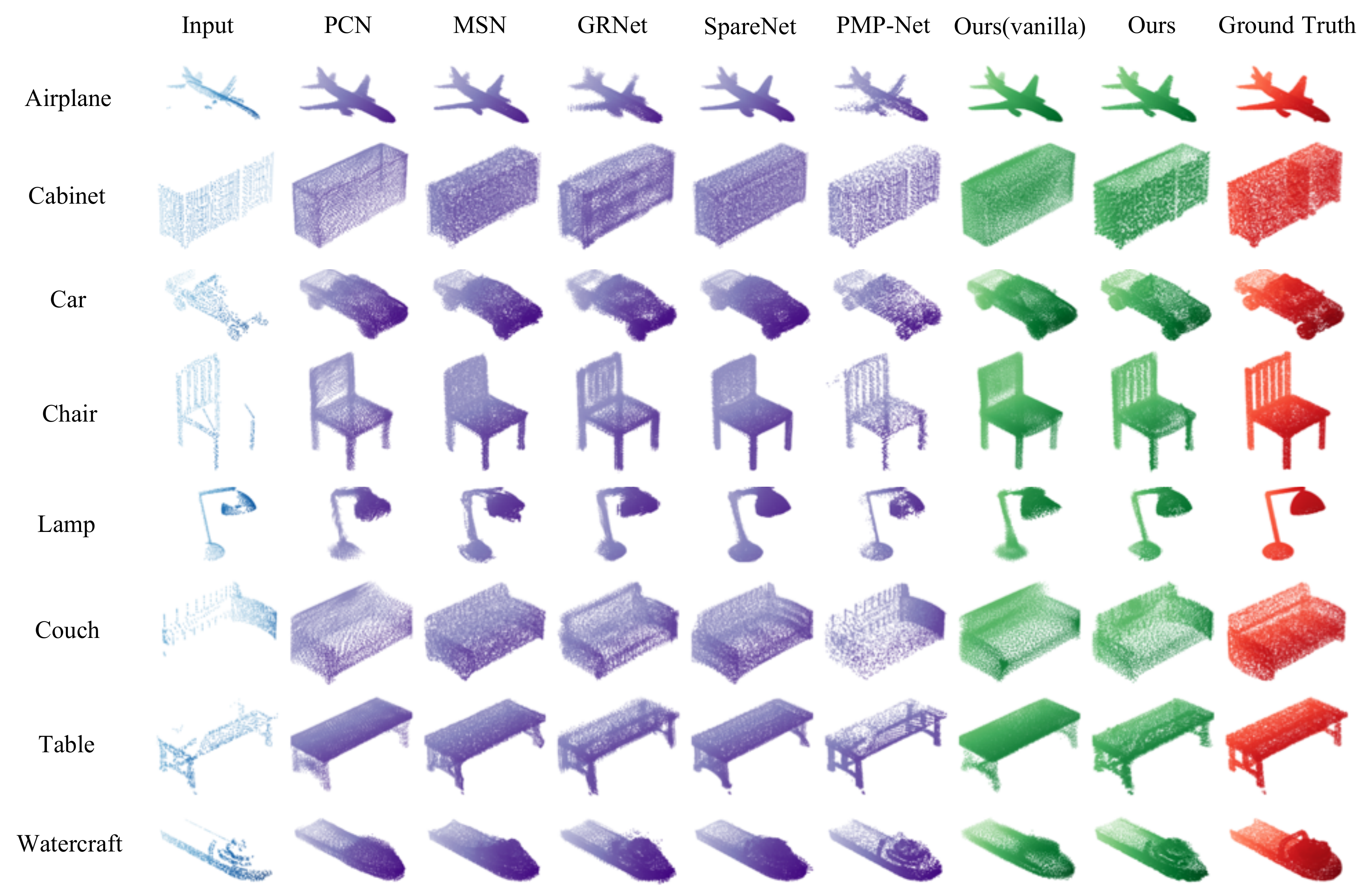}
    \caption{Visualized completion comparison on PCN dataset.}
    \label{fig:pcc_camparison}
\end{figure*}

AXform with one branch is compared with FC-based and folding-based methods on our sampled ShapeNet dataset. 
According to FoldingNet \cite{Yang_2018_CVPR}, 
the input feature dimension $K_1$ of the FC-based method and the folding-based method is set to 1024 and 512 respectively. 
The folding operation is performed in two rounds. 
To show that our method can still have better reconstruction results even network parameters are fewer than previous methods, 
we set $K_1=128$, $K_2=32$, and $N=128$. 
As shown in Table \ref{tab:reconstruction}, 
Chamfer Distance of our method is much better than previous two methods regardless of training on a single category or all three categories. 
As shown in Figure \ref{fig:recon_comparison}(a), 
it can be found that the reconstruction results of our method are closer to the ground truth. 
In addition, the reconstruction details of our method are better, such as the engines and tail of the airplane.

For AXform with K branches, we first do some ablation studies on the number of branches. 
As shown in Table \ref{tab:branches_ablation}, 
when the AXform in each branch is fixed, more branches will lead to better Chamfer Distance. 
However, the decline of Chamfer Distance decreases a lot when $K \geq 16$. 
When the network parameters are approximately kept unchanged, $K = 16$ obtains the optimal Chamfer Distance. 
Therefore, we choose $K = 16$ for the comparison experiments. 
Under the condition that network parameters are fewer than AtlasNet \cite{groueix2018}, 
as shown in Table \ref{tab:reconstruction}, our method can still achieve a lower Chamfer Distance. 
As shown in Figure \ref{fig:recon_comparison}(b), 
it can be found that the reconstruction results of our method are smoother and more even. 
For example, thin structures like chair legs are difficult to be deformed from 2d squares, so the result of AtlasNet is inferior to our method.

Figure \ref{fig:recon_loss} shows the loss curves of three methods during training. 
(a) and (b) represents the training process on a single airplane category and all three categories respectively. 
Since the loss value of the folding-based method is huge at the beginning, 
we use a dotted line to represent the huge value, 
which contributes to a clear comparison between the methods. 
It can be found that the convergence speed of AXform is significantly faster than the previous two methods, 
and our loss curve is relatively smoother.

AXform can be used on existing network architectures. 
For example, we replace the MLP decoder in l-GAN (CD) \cite{achlioptas2017latent_pc} to get better generation results. 
The generation set is the same size as the test set. 
To reduce the sampling bias of the evaluation metrics in Table \ref{tab:generation}, 
the process is repeated 3 times and reported the averages. 
It can be found that l-GAN-AXform achieves better JSD and 1-NNA and comparable MMD and COV. 
Compared with other metrics, \cite{Yang_2019_ICCV} points out that 1-NNA is better suited for evaluating generative models of point clouds. 
Therefore, AXform improves the point cloud generation model l-GAN (CD).

\subsection{Unsupervised Semantic Segmentation}

AXform with K branches can realize unsupervised semantic segmentation on the generated point clouds due to the property of space consistency. 
As shown in Figure \ref{fig:uss_results}(a), 
we give a realization workflow. 
When there are enough branches, 
the space consistency can be regarded as ``semantic consistency" which means each branch focuses on a certain semantics. 
Assume that the 16 branches are labeled 1-16. 
We take an airplane in the training set as a reference and find that branches 3, 4, and 7 generate the left-wing. 
Then, when reconstructing all the airplanes in the test set, branches 3, 4, and 7 will still generate the left-wing due to the property of space consistency. 
Finally, unsupervised semantic segmentation can be realized by merging the output of branches with the same semantics.

Figure \ref{fig:uss_results}(b) shows the unsupervised semantic segmentation results on three categories by using AXform with 16 branches. 
Here different colors represent different semantics. 
Airplane has three semantics of wing, body, and tail; Car has three semantics of wheel, body, and roof; Chair has three semantics of leg, surface, and back. 
It can be found that, except for some joints, the results are quite good. 
If more branches are used, the results can be more accurate.

\subsection{Point Cloud Completion}

We compare AXformNet with previous methods on the PCN dataset. 
Chamfer Distance and F-Score@1\% \cite{Tatarchenko_2019_CVPR} are used as evaluation metrics. 
In the approach section, 
we have pointed out that keeping the partial point cloud unchanged contributes to the metrics a lot. 
For a fair comparison, 
we run a vanilla version of AXformNet, 
which does not include the second refinement stage.

The metrics of SpareNet \cite{Xie_2021_CVPR} and PMP-Net \cite{wen2021pmp} 
in Table \ref{tab:pcc_comparison_cd} and Table \ref{tab:pcc_comparison_f1} are obtained from their given pretrained model 
and the others are obtained from the paper GRNet \cite{xie2020grnet}.
It can be found that AXformNet is the best in most categories and on average. 
The F-Score@1\% which is more convincing than Chamfer Distance is greatly improved. 
Figure \ref{fig:pcc_camparison} shows the visualized completion comparison on the PCN dataset. 
AXformNet can generate better complete point clouds than the other methods by using AXform. 
Since AXformNet(vanilla) is the ablation of the refinement module, 
here we only do ablation studies for the feature mapping module. 
As shown in Table \ref{tab:pcc_comparison_ablation}, 
the feature mapping module contributes to the improvement of the results. 
Without it, AXformNet can still behave well.

\section{Conclusion}

Since the transformation from latent features to point clouds has not been fully explored, 
we propose a novel attention-based method called AXform. 
It generates point clouds by weighting the points in an interim space 
and achieves better results than previous FC-based and folding-based methods. 
In addition, 
AXform has properties of self-clustering and space consistency when been expanded to multiple branches, 
which can be used for unsupervised semantic segmentation. 
We apply AXform to point cloud completion and it achieves state-of-the-art results on the PCN dataset.

\section{Acknowledgment}

This work was supported by National Natural Science Fund of China (62176064) and Zhejiang Lab (2019KD0AB06). Cheng Jin is the corresponding author.

% Use \bibliography{yourbibfile} instead or the References section will not appear in your paper
\bibliography{aaai22}

\begin{thebibliography}{48}
\providecommand{\natexlab}[1]{#1}

\bibitem[{Achlioptas et~al.(2017)Achlioptas, Diamanti, Mitliagkas, and
  Guibas}]{achlioptas2017latent_pc}
Achlioptas, P.; Diamanti, O.; Mitliagkas, I.; and Guibas, L.~J. 2017.
\newblock Learning Representations and Generative Models For 3D Point Clouds.
\newblock \emph{arXiv preprint arXiv:1707.02392}.

\bibitem[{Alliegro et~al.(2021)Alliegro, Valsesia, Fracastoro, Magli, and
  Tommasi}]{Alliegro_2021_CVPR}
Alliegro, A.; Valsesia, D.; Fracastoro, G.; Magli, E.; and Tommasi, T. 2021.
\newblock Denoise and Contrast for Category Agnostic Shape Completion.
\newblock In \emph{Proceedings of the IEEE/CVF Conference on Computer Vision
  and Pattern Recognition (CVPR)}, 4629--4638.

\bibitem[{Chang et~al.(2015)Chang, Funkhouser, Guibas, Hanrahan, Huang, Li,
  Savarese, Savva, Song, Su, Xiao, Yi, and Yu}]{Chang:2015:SAI}
Chang, A.~X.; Funkhouser, T.; Guibas, L.; Hanrahan, P.; Huang, Q.; Li, Z.;
  Savarese, S.; Savva, M.; Song, S.; Su, H.; Xiao, J.; Yi, L.; and Yu, F. 2015.
\newblock {ShapeNet}: An Information-Rich {3D} Model Repository.
\newblock Technical Report 1512.03012, arXiv preprint.

\bibitem[{Dai et~al.(2018)Dai, Ritchie, Bokeloh, Reed, Sturm, and
  Nie{\ss}ner}]{dai2018scancomplete}
Dai, A.; Ritchie, D.; Bokeloh, M.; Reed, S.; Sturm, J.; and Nie{\ss}ner, M.
  2018.
\newblock ScanComplete: Large-Scale Scene Completion and Semantic Segmentation
  for 3D Scans.
\newblock In \emph{Proc. Computer Vision and Pattern Recognition (CVPR), IEEE}.

\bibitem[{Fan, Su, and Guibas(2017)}]{Fan_2017_CVPR}
Fan, H.; Su, H.; and Guibas, L.~J. 2017.
\newblock A Point Set Generation Network for 3D Object Reconstruction From a
  Single Image.
\newblock In \emph{Proceedings of the IEEE Conference on Computer Vision and
  Pattern Recognition (CVPR)}.

\bibitem[{Fuchs et~al.(2020)Fuchs, Worrall, Fischer, and
  Welling}]{fuchs2020se3transformers}
Fuchs, F.~B.; Worrall, D.~E.; Fischer, V.; and Welling, M. 2020.
\newblock SE(3)-Transformers: 3D Roto-Translation Equivariant Attention
  Networks.
\newblock In \emph{Advances in Neural Information Processing Systems 34
  (NeurIPS)}.

\bibitem[{Groueix et~al.(2018)Groueix, Fisher, Kim, Russell, and
  Aubry}]{groueix2018}
Groueix, T.; Fisher, M.; Kim, V.~G.; Russell, B.; and Aubry, M. 2018.
\newblock {AtlasNet: A Papier-M\^ach\'e Approach to Learning 3D Surface
  Generation}.
\newblock In \emph{Proceedings IEEE Conf. on Computer Vision and Pattern
  Recognition (CVPR)}.

\bibitem[{Hu et~al.(2020)Hu, Yang, Xie, Rosa, Guo, Wang, Trigoni, and
  Markham}]{hu2019randla}
Hu, Q.; Yang, B.; Xie, L.; Rosa, S.; Guo, Y.; Wang, Z.; Trigoni, N.; and
  Markham, A. 2020.
\newblock RandLA-Net: Efficient Semantic Segmentation of Large-Scale Point
  Clouds.
\newblock \emph{Proceedings of the IEEE Conference on Computer Vision and
  Pattern Recognition}.

\bibitem[{Hu et~al.(2019)Hu, Han, Shrivastava, and Zwicker}]{Hu_2019_ICCV}
Hu, T.; Han, Z.; Shrivastava, A.; and Zwicker, M. 2019.
\newblock Render4Completion: Synthesizing Multi-View Depth Maps for 3D Shape
  Completion.
\newblock In \emph{Proceedings of the IEEE/CVF International Conference on
  Computer Vision (ICCV) Workshops}.

\bibitem[{Huang et~al.(2020)Huang, Yu, Xu, Ni, and Le}]{huang2020pfnet}
Huang, Z.; Yu, Y.; Xu, J.; Ni, F.; and Le, X. 2020.
\newblock PF-Net: Point Fractal Network for 3D Point Cloud Completion.
\newblock In \emph{IEEE Conference on Computer Vision and Pattern Recognition}.

\bibitem[{Hui et~al.(2020)Hui, Xu, Xie, Qian, and Yang}]{hui2020pdgn}
Hui, L.; Xu, R.; Xie, J.; Qian, J.; and Yang, J. 2020.
\newblock Progressive Point Cloud Deconvolution Generation Network.
\newblock In \emph{ECCV}.

\bibitem[{Kim et~al.(2020)Kim, Lee, Kang, Lee, and Kim}]{NEURIPS2020_bdbca288}
Kim, H.; Lee, H.; Kang, W.~H.; Lee, J.~Y.; and Kim, N.~S. 2020.
\newblock SoftFlow: Probabilistic Framework for Normalizing Flow on Manifolds.
\newblock In Larochelle, H.; Ranzato, M.; Hadsell, R.; Balcan, M.~F.; and Lin,
  H., eds., \emph{Advances in Neural Information Processing Systems},
  volume~33, 16388--16397. Curran Associates, Inc.

\bibitem[{Klokov, Boyer, and Verbeek(2020)}]{klokov20eccv}
Klokov, R.; Boyer, E.; and Verbeek, J. 2020.
\newblock Discrete Point Flow Networks for Efficient Point Cloud Generation.
\newblock In \emph{Proceedings of the 16th European Conference on Computer
  Vision (ECCV)}.

\bibitem[{Lee et~al.(2019)Lee, Lee, Kim, Kosiorek, Choi, and Teh}]{lee2019set}
Lee, J.; Lee, Y.; Kim, J.; Kosiorek, A.; Choi, S.; and Teh, Y.~W. 2019.
\newblock Set Transformer: A Framework for Attention-based
  Permutation-Invariant Neural Networks.
\newblock In \emph{Proceedings of the 36th International Conference on Machine
  Learning}, 3744--3753.

\bibitem[{Li et~al.(2018)Li, Zaheer, Zhang, Poczos, and
  Salakhutdinov}]{li2018point}
Li, C.-L.; Zaheer, M.; Zhang, Y.; Poczos, B.; and Salakhutdinov, R. 2018.
\newblock Point cloud gan.
\newblock \emph{arXiv preprint arXiv:1810.05795}.

\bibitem[{Li et~al.(2019)Li, Li, Fu, Cohen-Or, and Heng}]{li2019pugan}
Li, R.; Li, X.; Fu, C.-W.; Cohen-Or, D.; and Heng, P.-A. 2019.
\newblock PU-GAN: a Point Cloud Upsampling Adversarial Network.
\newblock In \emph{{IEEE} International Conference on Computer Vision
  ({ICCV})}.

\bibitem[{Liu et~al.(2019{\natexlab{a}})Liu, Sheng, Yang, Shao, and
  Hu}]{liu2019morphing}
Liu, M.; Sheng, L.; Yang, S.; Shao, J.; and Hu, S.-M. 2019{\natexlab{a}}.
\newblock Morphing and Sampling Network for Dense Point Cloud Completion.
\newblock \emph{arXiv preprint arXiv:1912.00280}.

\bibitem[{Liu et~al.(2019{\natexlab{b}})Liu, Han, Wen, Liu, and
  Zwicker}]{liu2019l2gautoencoder}
Liu, X.; Han, Z.; Wen, X.; Liu, Y.-S.; and Zwicker, M. 2019{\natexlab{b}}.
\newblock L2G Auto-encoder: Understanding Point Clouds by Local-to-Global
  Reconstruction with Hierarchical Self-Attention.
\newblock In \emph{Proceedings of the 27th ACM International Conference on
  Multimedia}.

\bibitem[{Luo and Hu(2021)}]{luo2021diffusion}
Luo, S.; and Hu, W. 2021.
\newblock Diffusion Probabilistic Models for 3D Point Cloud Generation.
\newblock In \emph{Proceedings of the IEEE/CVF Conference on Computer Vision
  and Pattern Recognition (CVPR)}.

\bibitem[{Pan(2020)}]{9093117}
Pan, L. 2020.
\newblock ECG: Edge-aware Point Cloud Completion with Graph Convolution.
\newblock \emph{IEEE Robotics and Automation Letters}, 5(3): 4392--4398.

\bibitem[{Pan et~al.(2021)Pan, Chen, Cai, Zhang, Zhao, Yi, and
  Liu}]{pan2021variational}
Pan, L.; Chen, X.; Cai, Z.; Zhang, J.; Zhao, H.; Yi, S.; and Liu, Z. 2021.
\newblock Variational Relational Point Completion Network.
\newblock \emph{arXiv preprint arXiv:2104.10154}.

\bibitem[{Qi et~al.(2017{\natexlab{a}})Qi, Su, Mo, and Guibas}]{qi2016pointnet}
Qi, C.~R.; Su, H.; Mo, K.; and Guibas, L.~J. 2017{\natexlab{a}}.
\newblock PointNet: Deep Learning on Point Sets for 3D Classification and
  Segmentation.
\newblock In \emph{IEEE Conference on Computer Vision and Pattern Recognition}.

\bibitem[{Qi et~al.(2017{\natexlab{b}})Qi, Yi, Su, and
  Guibas}]{qi2017pointnetplusplus}
Qi, C.~R.; Yi, L.; Su, H.; and Guibas, L.~J. 2017{\natexlab{b}}.
\newblock PointNet++: Deep Hierarchical Feature Learning on Point Sets in a
  Metric Space.
\newblock \emph{arXiv preprint arXiv:1706.02413}.

\bibitem[{Rui, Plinio, and Alexandre(2017)}]{visapp17}
Rui, F.; Plinio, M.; and Alexandre, B. 2017.
\newblock Automatic Object Shape Completion from 3D Point Clouds for Object
  Manipulation.
\newblock In \emph{Proceedings of the 12th International Joint Conference on
  Computer Vision, Imaging and Computer Graphics Theory and Applications -
  Volume 4: VISAPP, (VISIGRAPP 2017)}, 565--570. INSTICC, SciTePress.
\newblock ISBN 978-989-758-225-7.

\bibitem[{Sarmad, Lee, and Kim(2019)}]{Sarmad_2019_CVPR}
Sarmad, M.; Lee, H.~J.; and Kim, Y.~M. 2019.
\newblock RL-GAN-Net: A Reinforcement Learning Agent Controlled GAN Network for
  Real-Time Point Cloud Shape Completion.
\newblock In \emph{Proceedings of the IEEE/CVF Conference on Computer Vision
  and Pattern Recognition (CVPR)}.

\bibitem[{Shajahan, Nayel, and Muthuganapathy(2020)}]{8897042}
Shajahan, D.~A.; Nayel, V.; and Muthuganapathy, R. 2020.
\newblock Roof Classification From 3-D LiDAR Point Clouds Using Multiview CNN
  With Self-Attention.
\newblock \emph{IEEE Geoscience and Remote Sensing Letters}, 17(8): 1465--1469.

\bibitem[{Shu, Park, and Kwon(2019)}]{Shu_2019_ICCV}
Shu, D.~W.; Park, S.~W.; and Kwon, J. 2019.
\newblock 3D Point Cloud Generative Adversarial Network Based on Tree
  Structured Graph Convolutions.
\newblock In \emph{Proceedings of the IEEE/CVF International Conference on
  Computer Vision (ICCV)}.

\bibitem[{Sun et~al.(2020)Sun, Wang, Liu, Siegel, and Sarma}]{sun2018pointgrow}
Sun, Y.; Wang, Y.; Liu, Z.; Siegel, J.~E.; and Sarma, S.~E. 2020.
\newblock PointGrow: Autoregressively Learned Point Cloud Generation with
  Self-Attention.
\newblock In \emph{Winter Conference on Applications of Computer Vision}.

\bibitem[{Tatarchenko et~al.(2019)Tatarchenko, Richter, Ranftl, Li, Koltun, and
  Brox}]{Tatarchenko_2019_CVPR}
Tatarchenko, M.; Richter, S.~R.; Ranftl, R.; Li, Z.; Koltun, V.; and Brox, T.
  2019.
\newblock What Do Single-View 3D Reconstruction Networks Learn?
\newblock In \emph{Proceedings of the IEEE/CVF Conference on Computer Vision
  and Pattern Recognition (CVPR)}.

\bibitem[{Tchapmi et~al.(2019)Tchapmi, Kosaraju, Rezatofighi, Reid, and
  Savarese}]{topnet2019}
Tchapmi, L.~P.; Kosaraju, V.; Rezatofighi, S.~H.; Reid, I.; and Savarese, S.
  2019.
\newblock TopNet: Structural Point Cloud Decoder.
\newblock In \emph{IEEE Conference on Computer Vision and Pattern Recognition}.

\bibitem[{Valsesia, Fracastoro, and Magli(2019)}]{valsesia2018learning}
Valsesia, D.; Fracastoro, G.; and Magli, E. 2019.
\newblock Learning Localized Generative Models for 3D Point Clouds via Graph
  Convolution.
\newblock In \emph{International Conference on Learning Representations}.

\bibitem[{Varley et~al.(2017)Varley, DeChant, Richardson, Nair, Ruales, and
  Allen}]{varley2017shape}
Varley, J.; DeChant, C.; Richardson, A.; Nair, A.; Ruales, J.; and Allen, P.
  2017.
\newblock Shape Completion Enabled Robotic Grasping.
\newblock In \emph{Intelligent Robots and Systems (IROS), 2017 IEEE/RSJ
  International Conference on}. IEEE.

\bibitem[{Wang, Marcelo~H., and Lee(2020)}]{Wang_2020_CVPR}
Wang, X.; Marcelo~H., A.~J.; and Lee, G.~H. 2020.
\newblock Cascaded Refinement Network for Point Cloud Completion.
\newblock In \emph{Proceedings of the IEEE/CVF Conference on Computer Vision
  and Pattern Recognition (CVPR)}.

\bibitem[{Wang et~al.(2019)Wang, Sun, Liu, Sarma, Bronstein, and
  Solomon}]{dgcnn}
Wang, Y.; Sun, Y.; Liu, Z.; Sarma, S.~E.; Bronstein, M.~M.; and Solomon, J.~M.
  2019.
\newblock Dynamic Graph CNN for Learning on Point Clouds.
\newblock \emph{ACM Transactions on Graphics (TOG)}.

\bibitem[{Wang et~al.(2020)Wang, Tan, Navab, and
  Tombari}]{DBLP:journals/corr/abs-2008-07358}
Wang, Y.; Tan, D.~J.; Navab, N.; and Tombari, F. 2020.
\newblock SoftPoolNet: Shape Descriptor for Point Cloud Completion and
  Classification.
\newblock \emph{CoRR}, abs/2008.07358.

\bibitem[{Wen et~al.(2020)Wen, Li, Han, and Liu}]{Wen_2020_CVPR}
Wen, X.; Li, T.; Han, Z.; and Liu, Y.-S. 2020.
\newblock Point Cloud Completion by Skip-Attention Network With Hierarchical
  Folding.
\newblock In \emph{Proceedings of the IEEE/CVF Conference on Computer Vision
  and Pattern Recognition (CVPR)}.

\bibitem[{Wen et~al.(2021)Wen, Xiang, Han, Cao, Wan, Zheng, and
  Liu}]{wen2021pmp}
Wen, X.; Xiang, P.; Han, Z.; Cao, Y.-P.; Wan, P.; Zheng, W.; and Liu, Y.-S.
  2021.
\newblock PMP-Net: Point cloud completion by learning multi-step point moving
  paths.
\newblock In \emph{Proceedings of the IEEE Conference on Computer Vision and
  Pattern Recognition (CVPR)}.

\bibitem[{Xie et~al.(2021)Xie, Wang, Zhang, Yang, Chen, and
  Wen}]{Xie_2021_CVPR}
Xie, C.; Wang, C.; Zhang, B.; Yang, H.; Chen, D.; and Wen, F. 2021.
\newblock Style-Based Point Generator With Adversarial Rendering for Point
  Cloud Completion.
\newblock In \emph{Proceedings of the IEEE/CVF Conference on Computer Vision
  and Pattern Recognition (CVPR)}, 4619--4628.

\bibitem[{Xie et~al.(2020)Xie, Yao, Zhou, Mao, Zhang, and Sun}]{xie2020grnet}
Xie, H.; Yao, H.; Zhou, S.; Mao, J.; Zhang, S.; and Sun, W. 2020.
\newblock GRNet: Gridding Residual Network for Dense Point Cloud Completion.
\newblock In \emph{ECCV}.

\bibitem[{Yang et~al.(2019{\natexlab{a}})Yang, Huang, Hao, Liu, Belongie, and
  Hariharan}]{Yang_2019_ICCV}
Yang, G.; Huang, X.; Hao, Z.; Liu, M.-Y.; Belongie, S.; and Hariharan, B.
  2019{\natexlab{a}}.
\newblock PointFlow: 3D Point Cloud Generation With Continuous Normalizing
  Flows.
\newblock In \emph{Proceedings of the IEEE/CVF International Conference on
  Computer Vision (ICCV)}.

\bibitem[{Yang et~al.(2019{\natexlab{b}})Yang, Zhang, Ni, Li, Liu, Zhou, and
  Tian}]{Yang_2019_CVPR}
Yang, J.; Zhang, Q.; Ni, B.; Li, L.; Liu, J.; Zhou, M.; and Tian, Q.
  2019{\natexlab{b}}.
\newblock Modeling Point Clouds With Self-Attention and Gumbel Subset Sampling.
\newblock In \emph{Proceedings of the IEEE/CVF Conference on Computer Vision
  and Pattern Recognition (CVPR)}.

\bibitem[{Yang et~al.(2021)Yang, Wu, Zhang, and Jin}]{Yang_Wu_Zhang_Jin_2021}
Yang, X.; Wu, Y.; Zhang, K.; and Jin, C. 2021.
\newblock CPCGAN: A Controllable 3D Point Cloud Generative Adversarial Network
  with Semantic Label Generating.
\newblock \emph{Proceedings of the AAAI Conference on Artificial Intelligence},
  35(4): 3154--3162.

\bibitem[{Yang et~al.(2018)Yang, Feng, Shen, and Tian}]{Yang_2018_CVPR}
Yang, Y.; Feng, C.; Shen, Y.; and Tian, D. 2018.
\newblock FoldingNet: Point Cloud Auto-Encoder via Deep Grid Deformation.
\newblock In \emph{Proceedings of the IEEE Conference on Computer Vision and
  Pattern Recognition (CVPR)}.

\bibitem[{Yuan et~al.(2018)Yuan, Khot, Held, Mertz, and Hebert}]{yuan2018pcn}
Yuan, W.; Khot, T.; Held, D.; Mertz, C.; and Hebert, M. 2018.
\newblock PCN: Point Completion Network.
\newblock In \emph{International Conference on 3D Vision (3DV)}.

\bibitem[{Zamorski et~al.(2018)Zamorski, Zieba, Klukowski, Nowak, Kurach,
  Stokowiec, and Trzci{\'n}ski}]{zamorski2018adversarial}
Zamorski, M.; Zieba, M.; Klukowski, P.; Nowak, R.; Kurach, K.; Stokowiec, W.;
  and Trzci{\'n}ski, T. 2018.
\newblock Adversarial Autoencoders for Compact Representations of 3D Point
  Clouds.
\newblock \emph{arXiv preprint arXiv:1811.07605}.

\bibitem[{Zhang et~al.(2020)Zhang, Ma, Jiao, Liu, and Sun}]{ijcai2020-110}
Zhang, G.; Ma, Q.; Jiao, L.; Liu, F.; and Sun, Q. 2020.
\newblock AttAN: Attention Adversarial Networks for 3D Point Cloud Semantic
  Segmentation.
\newblock In Bessiere, C., ed., \emph{Proceedings of the Twenty-Ninth
  International Joint Conference on Artificial Intelligence, {IJCAI-20}},
  789--796. International Joint Conferences on Artificial Intelligence
  Organization.
\newblock Main track.

\bibitem[{Zhang, Yan, and Xiao(2020)}]{zhang2020eccv}
Zhang, W.; Yan, Q.; and Xiao, C. 2020.
\newblock Detail Preserved Point Cloud Completion via Separated Feature
  Aggregation.
\newblock In \emph{Computer Vision -- ECCV 2020}.

\bibitem[{Zhao et~al.(2020)Zhao, Jiang, Jia, Torr, and Koltun}]{zhao2020point}
Zhao, H.; Jiang, L.; Jia, J.; Torr, P.; and Koltun, V. 2020.
\newblock Point Transformer.
\newblock arXiv:2012.09164.

\end{thebibliography}

\end{document}